\documentclass{article}

\usepackage{arxiv}
\usepackage{colortbl}
\usepackage[utf8]{inputenc} % allow utf-8 input
\usepackage[T1]{fontenc}    % use 8-bit T1 fonts
\usepackage{hyperref}       % hyperlinks
\usepackage{url}            % simple URL typesetting
\usepackage{booktabs}       % professional-quality tables
\usepackage{amsfonts}       % blackboard math symbols
\usepackage{nicefrac}       % compact symbols for 1/2, etc.
\usepackage{microtype}      % microtypography
\usepackage{lipsum}
\usepackage{graphicx}
\graphicspath{ {./images/} }
\usepackage[table]{xcolor}
\definecolor{ZScolor}{RGB}{225,235,255}   % light blue
\definecolor{FTcolor}{RGB}{225,245,230}   % light green

\title{Evaluating Large Vision–language Models for Surgical Tool Detection}

\author{
 Nakul Poudel \\
  Center for Imaging Science\\
  Rochester Institute of Technology\\
  Rochester, NY 14623 \\
  \texttt{np1140@rit.edu} \\
   \And
 Richard Simon \\
  Biomedical Engineering\\
  Rochester Institute of Technology\\
  Rochester, NY 14623 \\
  \texttt{rasbme@rit.edu
} \\
  \And
 Cristian A. Linte \\
  Center for Imaging Science and Biomedical Engineering\\
  Rochester Institute of Technology\\
  Rochester, NY 14623 \\
  \texttt{calbme@rit.edu} \\
}

\begin{document}
\maketitle
\begin{abstract}

\textbf{Surgery is a highly complex process, and artificial intelligence has emerged as a transformative force in supporting surgical guidance and decision-making. However, the unimodal nature of most current AI systems limits their ability to achieve a holistic understanding of surgical workflows. This highlights the need for general-purpose surgical AI systems capable of comprehensively modeling the interrelated components of surgical scenes. Recent advances in large vision–language models that integrate multimodal data processing offer strong potential for modeling surgical tasks and providing human-like scene reasoning and understanding. Despite their promise, systematic investigations of VLMs in surgical applications remain limited. In this study, we evaluate the effectiveness of large VLMs for the fundamental surgical vision task of detecting surgical tools. Specifically, we investigate three state-of-the-art VLMs, Qwen2.5, LLaVA1.5, and InternVL3.5, on the GraSP robotic surgery dataset under both zero-shot and parameter-efficient LoRA fine-tuning settings. Our results demonstrate that Qwen2.5 consistently achieves superior detection performance in both configurations among the evaluated VLMs. Furthermore, compared with the open-set detection baseline Grounding DINO, Qwen2.5 exhibits stronger zero-shot generalization and comparable fine-tuned performance. Notably, Qwen2.5 shows superior instrument recognition, while Grounding DINO demonstrates stronger localization.
\linebreak \linebreak
}

\end{abstract}

% keywords can be removed

\section{Introduction}
Surgical environments are inherently complex, high-risk, and cognitively demanding. As procedures increasingly shift toward minimally invasive and robot-assisted techniques, the integration of Artificial Intelligence (AI) has emerged as a transformative force. Surgical AI systems provide critical intraoperative guidance and decision support, enhancing both surgical precision and patient safety~\cite{guo2025artificial}. However, most current surgical AI systems are unimodal and task-specific, with limited modeling of interactions among surgical tasks. Developing such capabilities is important for achieving comprehensive scene understanding, augmenting entire surgical workflows, and enabling future autonomous robotic assistants~\cite{khan2025surgical, schmidgall2024gp}.

Recent advances in vision-language models (VLMs) take multimodal input, i.e., images and rich textual descriptions, thereby enabling a more comprehensive understanding of the input data and modeling interactions among tasks. In the surgery domain, such models can encode information from surgical images and textual annotations, including the presence and type of instruments, anatomical structures, surgical actions, and their spatial relationships. This approach can enable a single model to perform multiple tasks, including instrument and tissue detection, surgical phase recognition, action recognition, step recognition, and critical view of safety. Moreover, the interdependencies among these tasks enable mutual reinforcement, leading to a more holistic understanding of surgical scenes. 

% Furthermore, these models enable surgeons to provide input instructions and receive results in natural language. 

% Consequently, VLMs represent a promising foundation for next-generation surgical AI systems that function as generalist agents capable of holistic surgical scene understanding, interactive assistance, and decision support.
Despite their promise, the effectiveness of VLMs in surgical environments depends on their visual perception capability. Errors in early visual perception can propagate to higher-level reasoning modules, potentially leading to unsafe or misleading predictions.  Among all surgical tasks, surgical tool detection and localization can be considered the fundamental component, as they directly support downstream tasks such as action recognition, phase recognition, and skill assessment. However, surgical scenes present substantial visual challenges, including occlusion, blood, smoke, motion blur, illumination variation, and poor contrast, which significantly degrade visual perception performance.

% Visual perception can be considered a fundamental surgical task. The accuracy of subsequent complex surgical tasks, such as surgical action and phase recognition, triplet recognition, and skill assessment, depends on the accuracy of visual perception. Conditions such as occlusion, blood, smoke, poor contrast, motion blur, and illumination impede the model's ability to achieve accurate visual perception. 

The application of VLMs in the surgical domain remains in its early stages. Several studies~\cite{schmidgall2024gp, zeng2025surgvlm} have developed surgery-specific VLMs by constructing large-scale surgical knowledge datasets, demonstrating encouraging results in multimodal reasoning. Other works~\cite{stueker2025vision, haider2025use} have explored the use of general-purpose VLMs for surgical instrument classification, showing promising potential. However, to the best of our knowledge, there is no systematic study that evaluates how general-purpose VLMs perform on the fundamental visual perception task of surgical instrument detection, particularly under both zero-shot and fine-tuned settings.

In this study, we present the comprehensive evaluation of three state-of-the-art large general-purpose vision-language models for surgical instrument detection. We investigate two practical deployment scenarios: (1) a zero-shot setting, in which off-the-shelf VLMs are directly prompted to classify and localize surgical instruments without any task-specific training, and (2) a fine-tuned setting, in which models are adapted to the surgical tool detection task using parameter-efficient LoRA fine-tuning. Through extensive experiments, we analyze the strengths and limitations of general-purpose VLMs for surgical visual perception and examine their potential applicability to future surgical AI systems.

\section{METHODOLOGY}

\subsection{Problem Definition}
Let $I \in \mathbb{R}^{H \times W \times 3}$ denote an input surgical image. The goal is to predict a set of surgical instrument instances
$\mathcal{O} = \{(c_i, b_i)\}_{i=1}^{N}$, where $c_i$ represents the instrument category, $b_i$ denotes the corresponding bounding box, and $N$ is the total number of instruments present in the image. We evaluate this task using three large VLMs, specifically Qwen2.5-7B~\cite{bai2025qwen2}, LLaVA1.5-7B~\cite{liu2023visual}, and InternVL3.5-8B~\cite{wang2025internvl3} across both zero-shot and fine-tuned settings.

\subsection{Dataset}
We used frames from the GraSP dataset~\cite{AYOBI2025103726}, extracted at $1$ fps from $13$ Robot-Assisted Radical Prostatectomy videos. Following the original dataset splits, we used frames extracted from five videos for testing, and the remaining videos for fine-tuning the models. The frames are annotated at $35$ frames interval, yielding $1125$ frames for testing and $2324$ frames for fine-tuning. There are a total of $7$ instrument categories, with each frame containing as few as $1$ and as many as $5$ instances. The distribution of instrument instances per instrument category is summarized in Table~\ref{tab:instrument_distribution}.

\begin{table}[h]
\centering
\small
\caption{Class-wise distribution of instrument instances in the training and test sets.}
\label{tab:instrument_distribution}
\begin{tabular}{lrrr}
\toprule
\textbf{Instrument Category} & \textbf{Train} & \textbf{Test} & \textbf{Total} \\
\midrule
Bipolar Forcep (BF)            & 1,694 & 809  & 2,503 \\
Prograsp Forcep (PF)           & 741   & 330  & 1,071 \\
Large Needle Driver (LND)      & 896   & 449  & 1,345 \\
Monopolar Curved Scissor (MCS) & 1,765 & 844  & 2,609 \\
Suction Instrument (SI)        & 816   & 310  & 1,126 \\
Clip Applier (CA)              & 64    & 38   & 102 \\
Laparoscopic Grasper (LG)      & 194   & 81   & 275 \\
\midrule
\textbf{All Instruments}       & \textbf{6,170} & \textbf{2,861} & \textbf{9,031} \\
\bottomrule
\end{tabular}
\end{table}

\subsection{Experimental Setup}
We experimented with three VLMs for the surgical tool detection task on the GraSP dataset. We imposed two settings—zero-shot and fine-tuned. For zero-shot inference, each model was prompted as-is, using its original pretrained weights to predict the instrument category and its corresponding bounding box on the test set. For the fine-tuned setting, we first adapted each model using supervised finetuning
with Rank-8 LoRA adaptation~\cite{hu2022lora} for $5$ epochs, a batch size of $4$, a gradient accumulation step of $4$, and a learning rate of $1\times e-4$ on the training set, and evaluated on the test set. The prompts used to query the model for fine-tuning and zero-shot inference are presented in Table~\ref{tab:prompts}. The prompt for zero-shot inference includes a list of potential instrument categories to guide the model. However, after fine-tuning, the model has already learned the instrument categories directly from the training data, and therefore, no such guidance is required. All the VLM experiments were performed using the Swift framework \footnote[4]{\url{https://github.com/modelscope/ms-swift}}  on an NVIDIA A100 (40GB) GPU~\cite{https://doi.org/10.34788/0s3g-qd15}. As a baseline, Grounding DINO~\cite{liu2024grounding} was fine-tuned for 15 epochs with a batch size of 4 and a learning rate of $1\times e-4$ using the Open-GroundingDino framework\footnote[5]{\url{https://github.com/longzw1997/Open-GroundingDino}}. During inference, a bounding box confidence threshold of 0.28 was applied to filter predicted bounding boxes.
\begin{table}[h]
\caption{Prompts used for zero-shot and fine-tuned VLM inference.}
\label{tab:prompts}
\centering
\small
\renewcommand{\arraystretch}{1.1}
\begin{tabular}{lp{12cm}}
\toprule
\textbf{Setting} & \textbf{Prompt} \\
\hline
Zero-shot & \texttt{<image>}From the following list of surgical instruments: [Bipolar Forcep, Prograsp Forcep, Large Needle Driver, Monopolar Curved Scissor, Suction Instrument, Clip Applier, Laparoscopic Grasper], identify which instruments are present in the given image. Also, give bounding box coordinates for them. Return your answer as a JSON object. \\
\hline
Fine-tuned & \texttt{<image>}Detect surgical instruments in the image. \\
\bottomrule
\end{tabular}
\end{table}
\begin{figure}[h]
    \centering
    \includegraphics[width=0.8\textwidth]{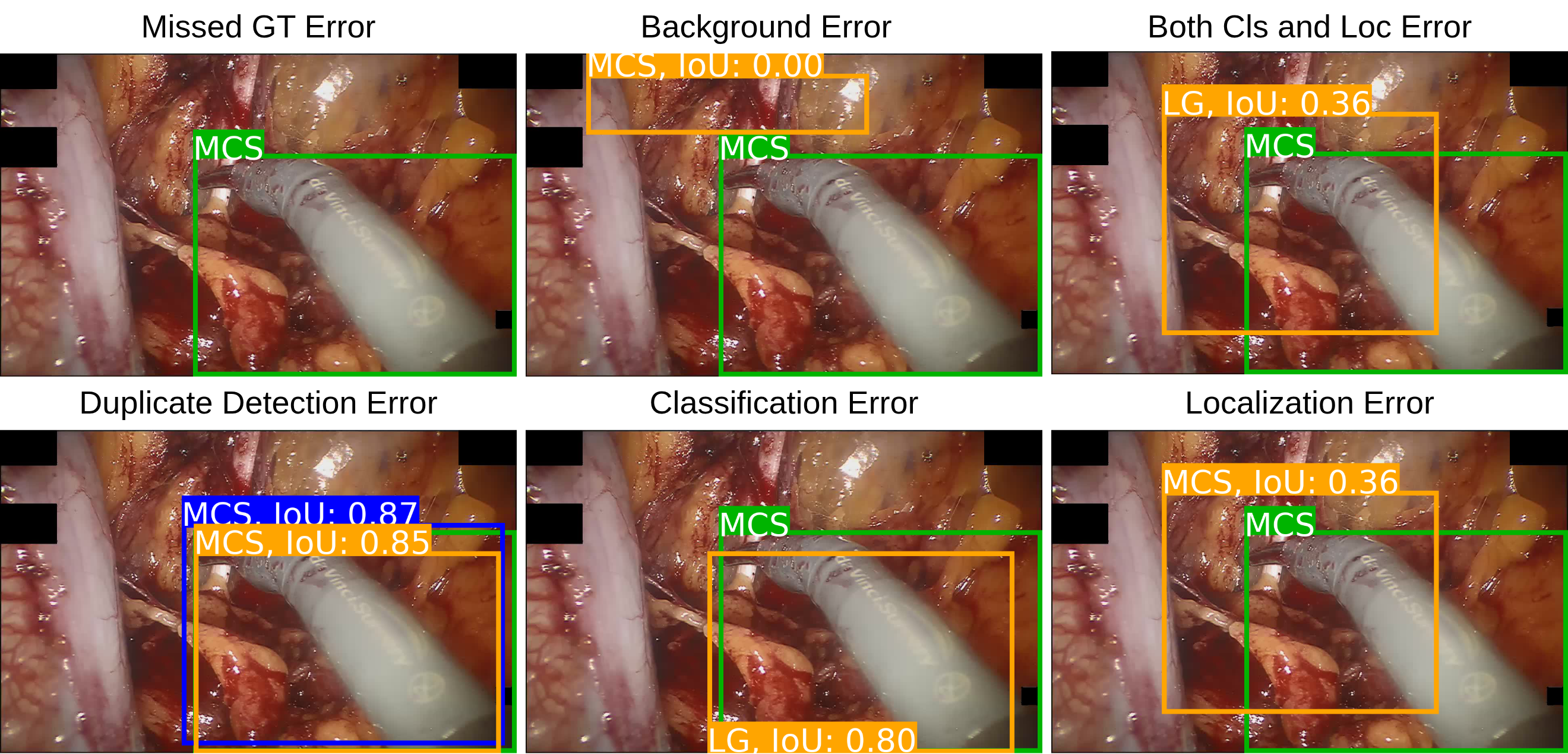}
    \caption{Illustration of error types. Green boxes indicate ground truth, and orange boxes indicate predictions corresponding to a specific error category. Shorthand instrument names are shown, along with the Intersection over Union (IoU) between the prediction and the ground truth box.}
    \label{fig:errors}
\end{figure}

\subsection{Evaluation Metrics}
The standard evaluation metric for object detection is mean Average Precision (mAP), which requires a confidence score for each predicted bounding box. In our VLM-based detection setting, the models generate bounding boxes and class labels through prompt-based inference, but do not provide confidence scores for each prediction. As a result, conventional mAP computation is not directly applicable. We therefore employ the TIDE framework~\cite{bolya2020tide} to evaluate detection performance in a confidence-independent manner by decomposing errors into multiple interpretable categories. This error-level analysis also provides more detailed insight into model behavior than a single aggregate performance score. 

The framework categorizes errors into six types, outlined below. Figure~\ref{fig:errors} illustrates representative examples of each category. We report the instance counts for the error categories. Specifically, the Missed GT category counts ground-truth instances, whereas all other categories count predicted instances. Here, $IoU_p$ denotes the Intersection over Union between a predicted and ground-truth bounding box, $t_f$ represents the foreground IoU threshold, and $t_b$ denotes the background IoU threshold, which are set to $0.5$ and $0.1$, respectively.

\begin{enumerate}
    \item \textbf{Missed GT Error (Miss):} A ground-truth object is not matched with any prediction.
    \item \textbf{Background Error (Bkg):} The prediction does not match any ground-truth object, i.e., $IoU_p \le t_b$.
    \item \textbf{Both Classification and Localization Error (Cls and Loc):} The prediction has both incorrect class assignment and insufficient localization accuracy, with $t_b \le IoU_p \le t_f$.
    \item \textbf{Duplicate Detection Error (Dup):} Multiple predictions correspond to the same ground-truth object, where only one prediction is considered correct, and the rest are counted as duplicates.

    \item \textbf{Classification Error (Cls):} A prediction is matched to a ground-truth object with $IoU_p \ge t_f$, but is assigned an incorrect class label.
    \item \textbf{Localization Error (Loc):} The predicted class is correct, but the bounding box localization is inaccurate, i.e., $t_b \le IoU_p \le t_f$.

\end{enumerate}

\section{RESULTS}
\begin{figure*}[t]

    \centering
    \includegraphics[width=\textwidth]{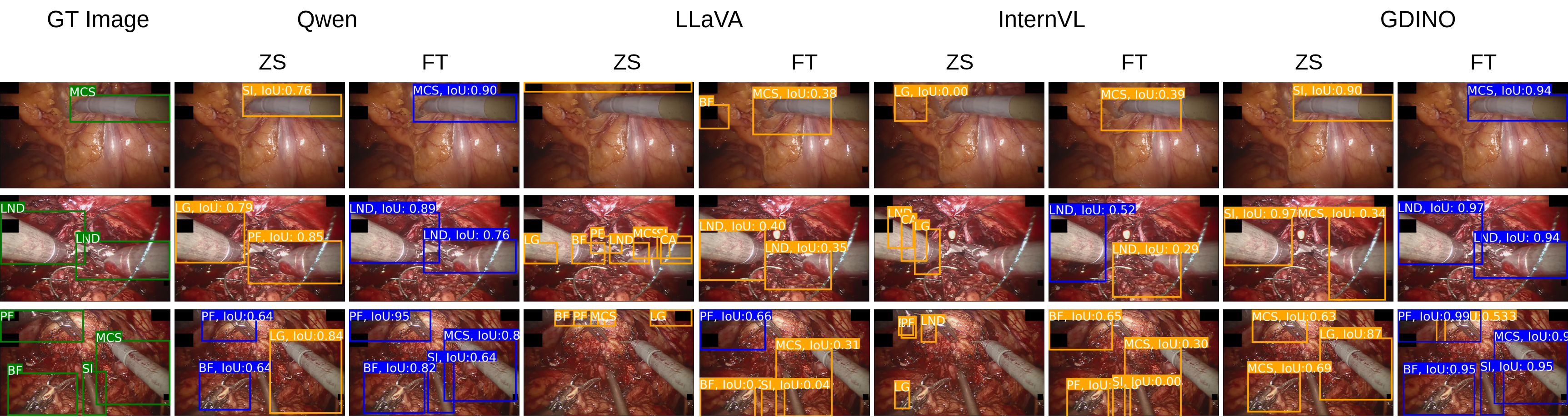}
    \caption{Qualitative comparison of surgical instrument detection across vision language models. Predictions from Qwen, LLaVA, and InternVL are shown alongside the GDINO baseline under both zero-shot (ZS) and fine-tuned (FT) settings. The green bounding box indicates ground truth, the blue bounding box indicates a correct detection, and the orange bounding box indicates an incorrect detection. Bounding boxes are annotated with abbreviated instrument class labels and, where applicable, their Intersection over Union (IoU) scores relative to the ground-truth (GT) annotations.}
    \label{fig:op}
\end{figure*}

\begin{table*}[h]
\caption{Error counts (number of instances) under different categories for three VLMs in zero-shot (ZS) and fine-tuned (FT) settings. GDINO is the baseline. For LLaVA and InternVL, individual classification and localization error counts are omitted in the ZS setting as the predictions are highly inconsistent with the ground truths. Lower errors are highlighted. }
\label{tab:vlm_errors}
\centering
\small
\resizebox{0.75\textwidth}{!}{%
\begin{tabular}{l
>{\columncolor{ZScolor}}c
>{\columncolor{FTcolor}}c
c c c c
>{\columncolor{ZScolor}}c
>{\columncolor{FTcolor}}c}
\toprule
\textbf{Errors} 
& \multicolumn{2}{c}{\textbf{Qwen}} 
& \multicolumn{2}{c}{\textbf{LLaVA}} 
& \multicolumn{2}{c}{\textbf{InternVL}} 
& \multicolumn{2}{c}{\textbf{GDINO (Baseline)}} \\
 & \textbf{ZS} & \textbf{FT} 
 & \textbf{ZS} & \textbf{FT} 
 & \textbf{ZS} & \textbf{FT} 
 & \textbf{ZS} & \textbf{FT} \\
\midrule
Missed GT     & $\mathbf{1,170}$& $\mathbf{329}$ & $1,723$ & $1,033$ & $2,623$ & $850$ & ${1,365}$& ${351}$ \\
Background    & ${337}$ & ${133}$ & $834$ & $573$ & $1,577$ & $582$  & $\mathbf{272}$ & $\mathbf{49}$ \\
Cls and Loc   & $\mathbf{471}$ & ${81}$  & $4,220$ & $329$  & $1,165$ & $289$ & ${673}$ & $\mathbf{17}$ \\
Duplicate     & $\mathbf{0}$   & $\mathbf{0}$   & $0$ & $0$ & $0$ & $0$ & ${2}$   & ${79}$ \\
Classification & $\mathbf{1,313}$ & $\mathbf{249}$ & $-$  & $54$ & $-$ & $54$ & ${1,516}$ & ${299}$ \\
Localization  & ${229}$ & ${140}$ & $-$ & $1,468$  & $-$ & $1,576$ & $\mathbf{171}$ & $\mathbf{56}$ \\
\bottomrule
\end{tabular}%
}
\end{table*}
The detection performance of three VLMs across six distinct error categories in both zero-shot and fine-tuned settings is summarized in Table~\ref{tab:vlm_errors}, and the visualization is presented in Figure~\ref{fig:op}. Grounding DINO, an open-set object detection model, serves as the baseline.

Among the tested VLMs, Qwen demonstrated strong detection performance in both zero-shot and fine-tuned settings. In the zero-shot setting, LLaVA predicted all seven instrument categories for every image, regardless of the number of instruments actually present, leading to repetitive, randomly placed bounding boxes and resulting in $4,220$ predicted instances counted as combined classification and localization error. InternVL, in contrast, produced only a limited number of predictions per image; however, most did not match the ground truth, resulting in $2,623$ missing instances out of $2,861$ ground-truth instances. Both models often misclassified the background as surgical instruments.

Under the zero-shot setting, the Qwen model detects instruments more effectively than the baseline, but is also more prone to misclassifying background regions as instruments. Despite this, Qwen exhibits a lower combined classification and localization error. Further category-level analysis shows that Qwen achieves lower classification error, i.e., better recognition of instrument categories, whereas the baseline demonstrates stronger localization performance. \

Fine-tuning substantially reduces errors across all categories for all models. After adaptation to the dataset, the baseline surpasses the Qwen in the combined classification and localization category. For the remaining error categories, the performance trend remains consistent with the zero-shot setting: the baseline continues to perform better in localization and background error categories, whereas the Qwen achieves superior performance in classification, duplicate detection, and missing ground-truth error categories. Notably, after fine-tuning, the baseline becomes more prone to duplicate predictions, whereas no such duplicates are observed with the Qwen model. Moreover, while Qwen's and GDINO's misdetections are due to misclassification, the misdetections of LLaVA and InternVL are due to poor localization.

Figure~\ref{fig:op} presents a quantitative comparison of detection results across different models. Consistent with our quantitative analysis, the detection strength of Qwen and GDINO is clearly evident. GDINO demonstrates superior localization performance, achieving higher IoU in most instances. The zero-shot behavior of LLaVA is also illustrated: in the first example, it predicts identical bounding boxes for all instrument categories; in the second example, it produces random bounding boxes for all categories; lastly, in the third example, it exhibits a mixture of both behaviors. Fine-tuning improves performance for LLaVA and InternVL, but these gains are insufficient to achieve correct detection.

\section{DISCUSSION AND CONCLUSIONS}
In this research, we evaluated three VLMs for surgical tool detection. The Qwen model demonstrated promising detection performance, while other VLMs struggled to correctly localize the tools, though they showed better tool classification after fine-tuning. We further compared the VLMs with an open-set detection-only baseline, GDINO. In the zero-shot setting, Qwen outperformed GDINO in four of the six instrument categories. After fine-tuning, the two models showed complementary strengths, each outperforming the other in three of the six categories. Overall, Qwen achieved superior classification performance, whereas GDINO exhibited stronger localization accuracy. This behavior is intuitive: GDINO is a detection-focused model trained on large-scale natural image datasets, which enables effective transfer of spatial localization to surgical scenes. However, due to the semantic gap in medical instrument domains, it often fails to assign correct class labels. In contrast, Qwen benefits from large-scale image–text pretraining, which facilitates more effective semantic transfer and improves classification of surgical tools. We also tested YOLO-World~\cite{Cheng2024YOLOWorld} as another open-vocabulary detection model; however, reasonable thresholds often yielded no predictions, whereas lower thresholds caused excessive false detections, making threshold selection challenging.

Overall, these encouraging results from VLMs, particularly Qwen, highlight the potential to develop general-purpose surgical VLMs capable of comprehensive analysis of surgical workflows. Future research could extend this work by leveraging Qwen across multiple surgical tasks, such as phase, action, and step recognition. 

\bibliographystyle{unsrt}  
\bibliography{references}  %%% Remove comment to use the external .bib file (using bibtex).


\end{document}